%% file: ms.tex
\documentclass[10pt,twocolumn,letterpaper]{article}

\usepackage{titlesec}
\titlespacing*{\paragraph}{0pt}{.5ex plus .5ex minus .5ex}{1em}
\titlespacing*{\subsubsection}{0pt}{6pt plus 2pt minus 2pt}{4pt plus 2pt minus 2pt}

\usepackage{iccv}
\usepackage{times}
\usepackage{epsfig}
\usepackage{graphicx, animate}
\usepackage{amsmath}
\usepackage{amssymb}

\usepackage{amsmath}
\DeclareMathOperator*{\argmax}{\arg\!\max}
\DeclareMathOperator*{\argmin}{\arg\!\min}

\usepackage{algorithm}
\usepackage{algpseudocode}
\usepackage{booktabs}
\usepackage{multirow}
\usepackage{subcaption}
\usepackage{tabularx}
\usepackage[dvipsnames]{xcolor}
\usepackage{subfiles}
\usepackage[format=plain,labelformat=simple,labelsep=period,font=small,compatibility=false]{caption}
\usepackage{enumitem}
\usepackage{dirtytalk}
\usepackage{appendix}

\usepackage[accsupp]{axessibility}  

\usepackage[pagebackref=true,breaklinks=true,letterpaper=true,colorlinks,bookmarks=false]{hyperref}

\newcommand{\modelname}{\textbf{{XMem++}}}
\newcommand{\datasetname}{\textbf{{PUMaVOS}}}
\newcommand{\good}[1]{#1}

\iccvfinalcopy 

\def\iccvPaperID{12179} 

\ificcvfinal\fi

\begin{document}

\title{XMem++: Production-level Video Segmentation From Few Annotated Frames}

\author{
    Maksym Bekuzarov$^{1*}$, Ariana Bermudez$^{1*}$\\
    {\tt\small \{maksym.bekuzarov,ariana.venegas\}@mbzuai.ac.ae}
    \and
    Joon-Young Lee $^{3}$ \\
    {\tt\small  jolee@adobe.com}
    \and
    Hao Li $^{1,2}$ \\
    {\tt\small hao@hao-li.com}
    \and 
        $^1$MBZUAI
    \and  
        $^2$Pinscreen
    \and 
        $^3$Adobe Research
}

\twocolumn[{
\maketitle\thispagestyle{empty}
\vspace{-6mm}
\centering
\includegraphics[width=0.89\linewidth]{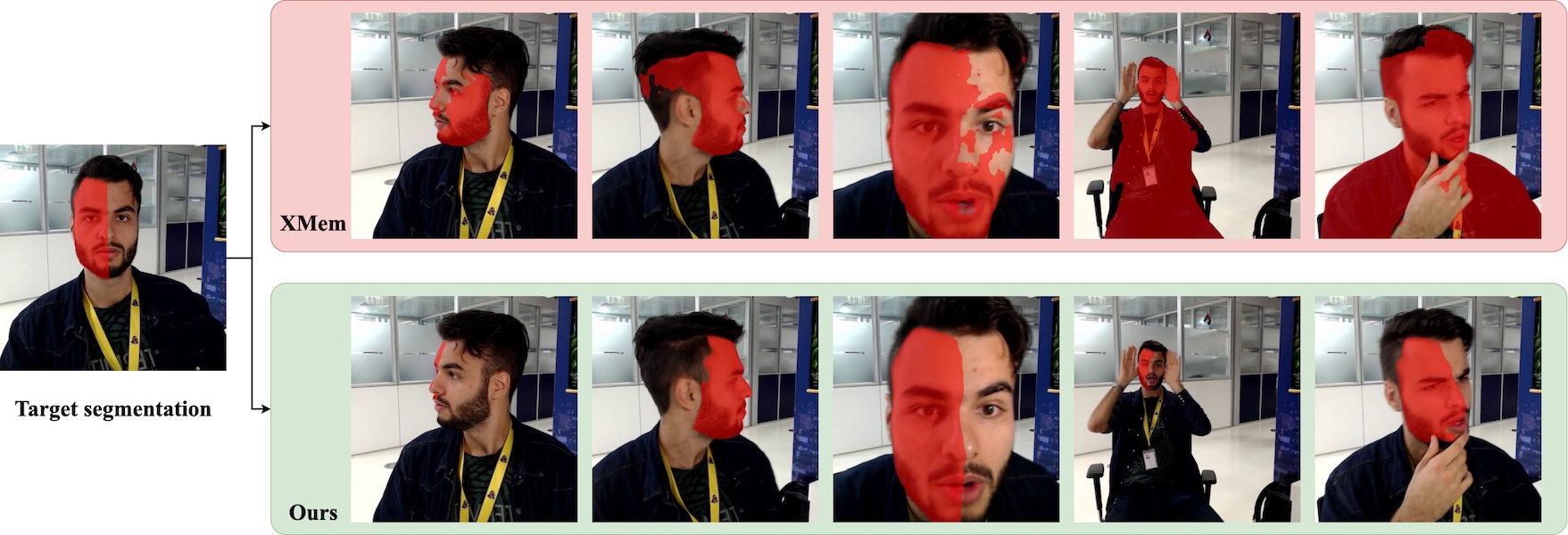}
\vspace{0mm}
\captionof{figure}{\small A demonstration of partial region segmentation (left half of the face) for extreme poses. We compare our method with the current SOTA, XMem~\cite{cheng2022xmem}, for a 1 min video (1800 frames), with only 6 frames (0.33\%) annotated. No retraining or fine-tuning required. } \label{fig:teaser}
\vspace{3mm}
}]

\def\thefootnote{*}\footnotetext{These authors contributed equally to the work}

\subfile{sections/abstract}

\subfile{sections/introduction}
\subfile{sections/relatedworks}
\subfile{sections/method}
\subfile{sections/experiments}

\subfile{sections/speed}
\subfile{sections/limitations}
\subfile{sections/conclusion}
{\small
\bibliographystyle{ieee_fullname}
\bibliography{ms}

}

\newpage
\appendixpage
\subfile{sections/appendix}

\end{document}

%% file: sections/abstract.tex
\begin{abstract}

Despite advancements in user-guided video segmentation, extracting complex objects consistently for highly complex scenes is still a labor-intensive task, especially for production. It is not uncommon that a majority of frames need to be annotated. We introduce a novel semi-supervised video object segmentation (SSVOS) model, \modelname{}, that improves existing memory-based models, with a permanent memory module. 
Most existing methods focus on single frame annotations, while our approach can effectively handle multiple user-selected frames with varying appearances of the same object or region. Our method can extract highly consistent results while keeping the required number of frame annotations low. 
We further introduce an iterative and attention-based frame suggestion mechanism, which computes the next best frame for annotation. Our method is real-time and does not require retraining after each user input. We also introduce a new dataset, \datasetname{}, which covers new challenging use cases not found in previous benchmarks. We demonstrate SOTA performance on challenging (partial and multi-class) segmentation scenarios as well as long videos, while ensuring significantly fewer frame annotations than any existing method. Project page: \href{https://max810.github.io/xmem2-project-page/}{https://max810.github.io/xmem2-project-page/}

\end{abstract}

%% file: sections/introduction.tex
\section{Introduction}

Video Object Segmentation (VOS)~\cite{VOS_survey} is a widely performed vision task with applications ranging from object recognition, scene understanding, medical imaging, to filter effects in video chats. While fully automated approaches based on pre-trained models for object segmentation are often desired, interactive user guidance is commonly practiced to annotate new training data or when precise rotoscoping is required for highly complex footages such as those found in visual effects. This is particularly the case when the videos have challenging lighting conditions and dynamic scenes, or when partial region segmentation is required. While automatic VOS methods are designed to segment complete objects with clear semantic outlines, interactive video object segmentation (IVOS) and semi-supervised video object segmentation (SSVOS) techniques~\cite{VOS_survey} are more flexible, and typically use a scribble or contour drawing interface for manual refinement such as those found in commercial software solutions such as Adobe After Effects and Nuke. Despite advancements in IVOS and SSVOS techniques, rotoscoping in film production is still a highly labor-intensive task, and often requires nearly every frame of a shot to be annotated and refined \cite{adoberotoscope2017}.

State-of-the-art IVOS and SSVOS techniques use memory-based models~\cite{STM} and have shown impressive segmentation results on complex scenes based on user-provided mask annotations, but they are often designed to improve single annotation performances~\cite{STM, AOT, DeAOT, STCN, cheng2022xmem}  and are still not suitable for production use cases. In particular, they tend to over-segment known semantic outlines (person, hair, faces, entire objects) and fail on partial regions (e.g., half of a person's face, a dog's tail), harsh lighting conditions such as shadows, and extreme object poses. As a result, inconsistent segmentations are obtained when only a single annotated frame is provided, due to the inherent ambiguity of the object's appearance, Especially when it varies too much due to large viewing angles and complex lighting conditions, which limits the scalability of these techniques, and it is often unclear which frame annotations to prioritize, especially for long sequences.

We propose a new SSVOS framework, \modelname{}, which uses a permanent memory module that stores all annotated frames and makes them available as references for all the frames in the video. While most SSVOS methods focus on single-frame mask annotations, our approach is designed to handle multiple frames that can be updated by the user iteratively with optimal frames being recommended by our system. While we adopt the cutting-edge architecture of XMem \cite{cheng2022xmem} as backbone, we show that our important modification enables accurate segmentation of challenging video objects (including partial regions) in complex scenes with significantly fewer annotated frames than existing methods. Our modification does not require any re-training or calibration and additionally shows improved temporal coherence in challenging scenarios (Fig. \ref{fig:teaser}, \ref{fig:xmempp_results}, \ref{fig:xmemppcomparison}). Our attention-based frame suggestion method predicts the next candidate frame for annotation based on previous labels while maximizing the diversity of frames being selected. Our system supports both sparse (scribbles) and dense (full masks) annotations and yields better quality scaling with more annotations provided than existing methods. The video segmentation performs in real-time, and frame annotations are instantly taken into account with a pre-trained network.





We further introduce a new dataset, \datasetname{} for benchmarking purposes, which includes new challenging scenes and use cases, including occlusion, partial segmentation, and object parts segmentation, where the annotation mask boundaries may not  correspond to visual cues.

We evaluate the performance of our algorithm both qualitatively and quantitatively on a wide range of complex video footages as well as existing datasets, and demonstrate SOTA segmentation results on complex scenes. In particular, we show examples where our method achieves higher accuracy and temporal coherence than existing methods with up to $5 \times$ fewer frame annotations and on $2 \times$ twice less on existing benchmarks (Section \ref{sec:results}).
We further demonstrate the effectiveness of our frame annotation candidate selection method by showing that it selects semantically meaningful frames, similar to those chosen by expert users. 
We make the following contributions: 
\begin{itemize}[noitemsep,topsep=0pt]
    \item We have introduced a new VOS model, \modelname{}, that uses a permanent memory module that effectively utilizes multiple frame annotations and produces temporally-smooth segmentation results without overfitting to common object cues. 
    \item We further propose the use of an attention-based similarity scoring algorithm that can take into account previously predicted frame annotations to suggest the next best frame for annotation.
    \item We present a new dataset, \datasetname{}, which contains long video sequences of complex scenes and non object-level segmentation cues, which cannot be found in existing datasets.
    \item We achieve SOTA performance on major benchmarks, with significantly fewer annotations, and showcase successful examples of complex multi-class and partial region segmentations that fail for existing techniques.
\end{itemize}

%% file: sections/relatedworks.tex
\section{Related Works}
\label{sec:related_works}
\paragraph{Video Object Segmentation.} 
A wide range of video object segmentation (VOS) methods have been introduced in the past decade~\cite{VOS_survey}, spanning a broad spectrum of computer vision applications, including visual effects. Many solutions have been deployed in established commercial video editing software such as Adobe After Effects and Nuke. While early VOS techniques were often based on classic optimization methods and graph representations~\cite{SeamSeg2014, VOSpredp2012}, recent ones are typically using deep neural networks.

Semi-supervised video object segmentation (SSVOS) aims at segmenting objects in a video using a frame of reference (usually the first \cite{OSVOS2017} but some models also support multiple annotations). To help users create the annotation, interactive VOS methods (IVOS) were introduced \cite{joon2019, MIVOS}, providing users a convenient way to create annotation masks commonly using scribbles and dots selection interface.  

To facilitate user annotations, some IVOS methods suggest a fine-tuning approach, which makes any iterative user interaction slow during inference as retraining is required~\cite{OSVOS2017, xiao2018monet}. Although more efficient alternatives like online adaptation were introduced, their output quality is generally poorer~\cite{NEURIPS2020_eosvos, Robinson_2020_CVPR, park2021learning, bhat2020learning}. 

Attention-based methods use different techniques such as similarity or template matching algorithms to dictate which frames need to be focussed on from a set of available frames (referred to as memories)~\cite{STM, duarte2019capsulevos, zhang2020transductive,huang2020fast, ge2021video}. 
Multiple authors have focused on facilitating the model to use local/pixel-to-pixel information which improves the quality of the masks using either kernels\cite{KMN2020}, optical flow\cite{xie2021efficient, yu2022batman}, transformers \cite{mao2021joint, lan2022learning, AOT, DeAOT, yu2022batman} or improvements to the spatial-temporal memory \cite{ wang2021swiftnet, xu2022reliable, MIVOS, STCN, AFB_UBR2020, li2020fast, QDMN2022, SWEM2022, li2022recurrent}. 

Most recently, XMem~\cite{cheng2022xmem} has proposed a resource-efficient method that compresses the feature memory and supports the usage of multiple annotated frames references. We base our approach on it, as the architecture is resource-efficient, quick, extendable, and demonstrates SOTA results on modern benchmarks.


\paragraph{Frame Annotation Candidate Selection.}

The task of annotation candidate prediction is finding a specific frame or set of frames for the user to annotate, in the first or consecutive interaction rounds that maximize the overall video segmentation quality (defined with metrics like \textit{Intersection over Union} (IoU) and \textit{F-score}). The authors of BubbleNets \cite{BubbleNets} propose a VOS architecture that simultaneously learns to predict the optimal candidate frame to annotate, by using uses a bubble-sort \cite{bubblesort} style algorithm to find the close-to-best candidate. The authors of IVOS-W \cite{IVOS-W} claim that annotating the frame with the lowest quality is suboptimal and introduce a Reinforcement Learning Agent to intelligently select the candidates based on the assessed quality of all frames. GIS-RAmap \cite{GISRAmap} introduces an end-2-end deep neural network, that operates on sparse user input (scribbles), and uses the R-attention mechanism to segment frames and directly predicts the best annotation candidates.  
\begin{figure}
\captionsetup[subfigure]{labelformat=empty}
    \centering
    \begin{subfigure}[b]{0.32\columnwidth}
        \centering
        \includegraphics[width=0.95\columnwidth]{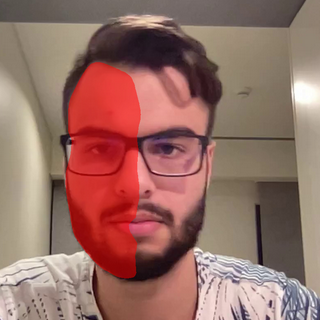}
        \caption{{\small Target, AQ = 97.0}}    
        \label{fig:mean and std of net14}
    \end{subfigure}
    \begin{subfigure}[b]{0.32\columnwidth}  
        \centering 
        \includegraphics[width=0.95\columnwidth]{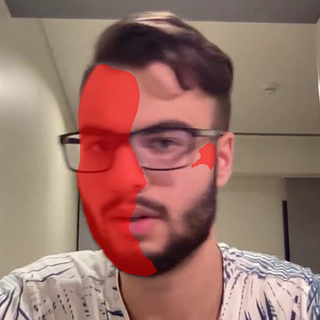}
        \caption[]{{\small Good VQ, AQ = 87.8}} 
        \label{fig:mean and std of net24}
    \end{subfigure}
    \begin{subfigure}[b]{0.32\columnwidth}   
        \centering 
        \includegraphics[width=0.95\columnwidth]{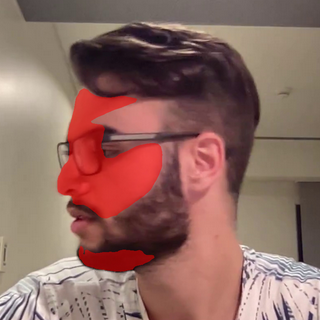}
        \caption[]{\small Poor VQ, AQ= \textbf{99.3}}    
         \label{fig:broken_quality_assessement}
    \end{subfigure}
    \hfill
    \vspace{-1mm}
        \caption{\small Assessed Quality ($AQ$) and Visual Quality ($VQ$) using quality assessement modele from IVOS-W \cite{IVOS-W}, demonstrated on one of the videos from \datasetname. \textbf{Left half of the face} is being segmented. \textit{Assessed quality does not correspond with visual quality for partial segmentation}.} 
    \vspace{-3mm}
    \label{fig:mean and std of nets}
\end{figure}
\begin{figure*}[h]
    \centering

    \includegraphics[width=1.0\textwidth]{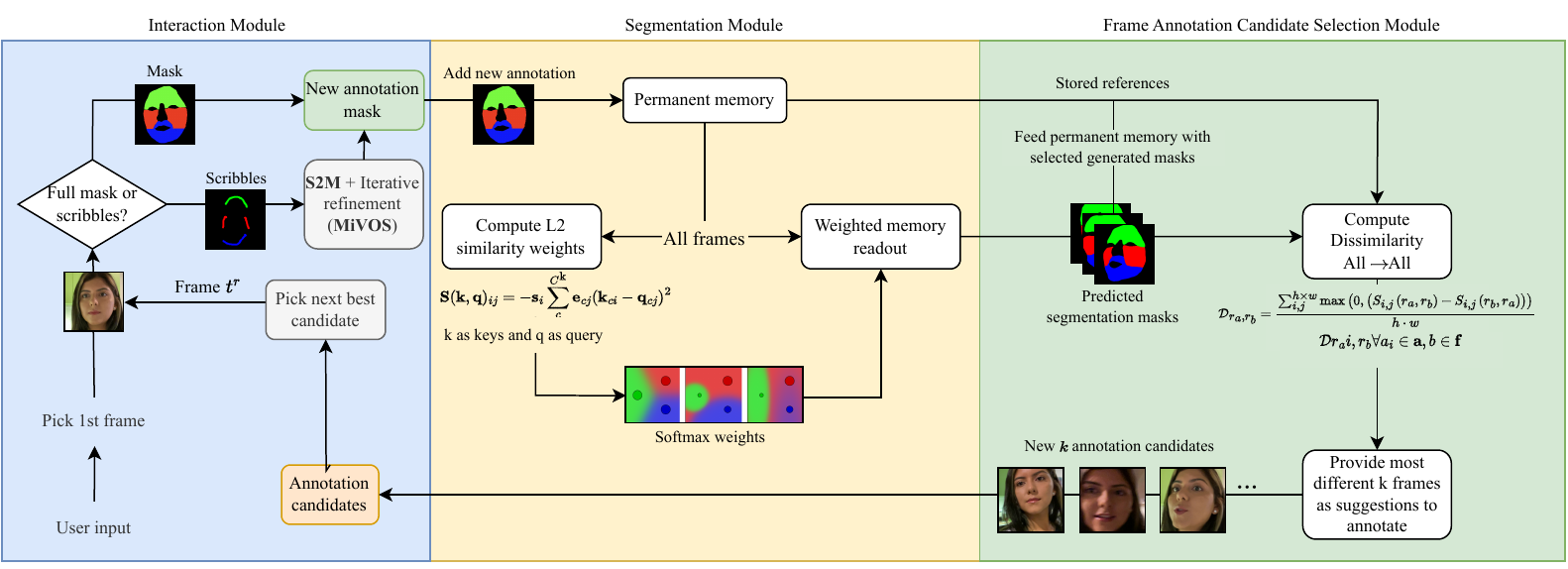}
   \vspace{-1mm}
    \caption{Interaction flow of our system. The user provides initial annotations, segmentation is performed, then using predicted masks new annotation candidates are chosen and given to the user. This loop is repeated until satisfactory segmentation quality is achieved.}
    \vspace{-2mm}
    \label{fig:xxxmem_flow}
\end{figure*}

These works exhibit the following limitations: 
\cite{BubbleNets,IVOS-W} work under the assumption that it is possible to directly estimate the segmentation quality or frame importance without explicit information of the target object, which makes them highly domain-dependent and does not hold for partial segmentation, as illustrated by Fig. \ref{fig:broken_quality_assessement}. \cite{BubbleNets} and \cite{GISRAmap} do not use annotation information when selecting annotation candidates, which limits their usefulness for partial segmentation, occlusions, or multiple similar objects in the scene (use-case illustrated in rows 1-2 in Fig. \ref{fig:frame_selector}).

\paragraph{Video Segmentation Datasets.}
Earlier datasets had annotations for videos without heavy occlusions or appearance changes \cite{ochs2013segmentation, jain2014supervoxel, fan2015jumpcut}. DAVIS \cite{DAVIS, DAVIS2017} became the benchmark dataset for the previous decade and is still being used because of the high resolution of the videos and the quality of the annotations. The benchmark released in 2016 had only single-object annotations, and the extended one in 2017 incorporated multiple-object annotations. For both datasets, their clips are 2-4 seconds long. Youtube-VOS \cite{YOUTUBE-VOS} presented a large dataset of around 4000 videos where each is 3-6 seconds long. Given the number of videos, they have a variety of categories and every 5-th frame is annotated. OVIS \cite{OVIS} is a dataset of severe occlusions, and is very challenging for the current VOS methods. MOSE \cite{MOSE} uses a subset of OVIS \cite{OVIS} as well as other videos, for a total of 2149 clips, from 5 to 60 seconds long, 12 on average. LVOS \cite{LVOS} dataset consists of 220 long videos, on average 115 seconds each. Most recently, BURST \cite{BURST} introduced a large dataset with almost 3000 videos, that can be used for tasks like VOS and Multi-Object Tracking and Segmentation. Even though there are resources with long videos and high-quality masks, many use cases from the industry are not reflected, such as partial objects, reflections, and segmentation targets without clear boundaries. 

%% file: sections/method.tex
\section{\modelname{}}
\subsection{Overview}

Fig. \ref{fig:xxxmem_flow} provides an overview of XMem++. Given a video, the end user first selects a frame $f_i$ they want to annotate, to give the model the template of the object(s) that will be segmented. The user then provides the annotations in the form of scribbles (that are then converted to a dense segmentation mask $m_i$) or the mask $m_i$ directly. The segmentation model (Section \ref{sec:permanent_memory}) then processes $m_i$, puts it into the permanent memory and segments the given target(s) in all other frames by predicting segmentation masks $\hat{\mathbf{m}}$. The annotation candidate selection algorithm (Section \ref{sec:frame_selection}) takes $m_i$ and $\hat{\mathbf{m}}$ and predicts the $k$ next best frames to annotate, in order of importance. The user then provides the annotations for some or all of them, and the process is repeated until the segmentation quality is satisfactory. The annotation candidate selection module takes into account all of the previously annotated frames, so it avoids selecting the frames that are similar to those already annotated.

\begin{figure*}[h]
    \centering
    \includegraphics[width=0.69\textwidth]{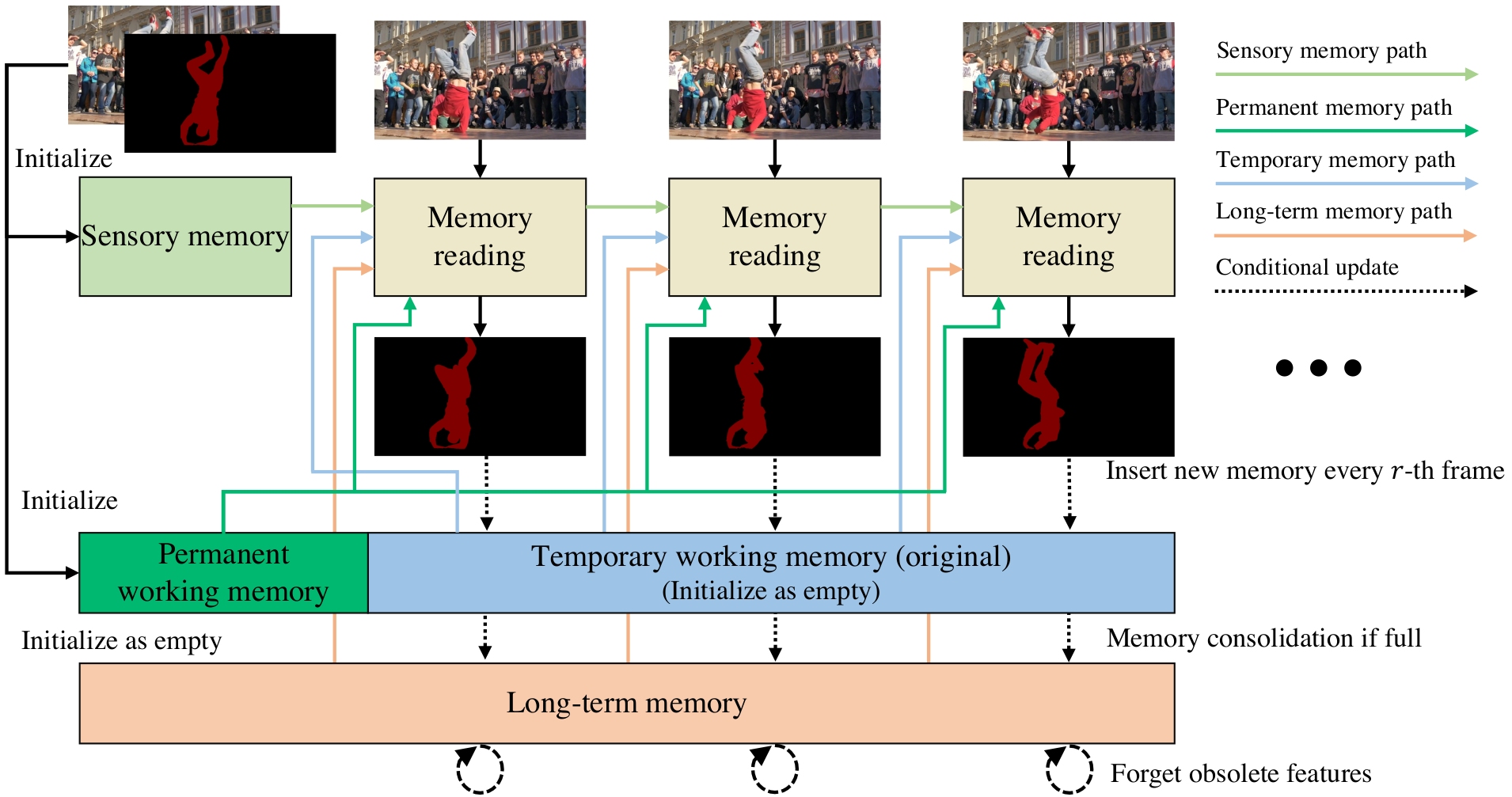}
   
\vspace{-3mm}
    \caption{\modelname{} architecture. The new permanent memory module is shown in {{dark-green}}.}
\vspace{-2mm}
    \label{fig:xxxmem_architecture}
\end{figure*}

The segmentation module is described in Fig. \ref{fig:xxxmem_architecture}. It is based on XMem architecture \cite{cheng2022xmem} and consists of a convolutional neural network with multiple parts and three types of separate memory modules. Given a sequence of frames $\mathbf{f}$ and at least one segmentation mask $m_i$ containing the target object(s), the mask is processed together with the corresponding frame $f_i$ by the model and stored in the permanent working memory as a reference for segmenting other frames. For every frame in the memory, two feature maps are extracted and saved - a smaller ``key", containing information about the whole frame, used for matching similar frames, and a corresponding larger ``value" with target-specific information, used in predicting the segmentation mask. When predicting the segmentation for a frame $f_j$, the model searches for similar frames in all three memory modules by calculating pixel-wise attention across stored ``keys" using a scaled $L_2$ similarity measure, aggregates the information from them and uses it to predict the segmentation for the frame $f_j$. The model also stores its own predictions in the temporary working memory modules and uses them together, usually every $n$-th frame. The long-term memory module was introduced in XMem. It limits memory usage and allows processing longer videos by frequently compressing and removing outdated features from temporary working memory. Sensory memory is a small module that captures motion information by processing differences between consecutive frames, also unchanged from \cite{cheng2022xmem}.   

\subsection{SSVOS with Permanent Memory Module}
\label{sec:permanent_memory}


Our main contribution is the introduction of a ``permanent" working memory module, that changes how the annotated frames are treated before and during model inference. In the original work, there was only temporary working memory, so the first frame was always annotated and permanently kept there for predicting segmentation masks for new frames. We define this frame as a ``ground-truth reference", meaning that the segmentation mask associated with it is 100\% accurate because it is provided by the user. All the other frames, which would be added to the working memory, were only added there temporarily and are likely to be moved to long-term memory eventually, especially for longer videos. We define these frames as ``imperfect references", as the associated segmentation mask is predicted by the model, thus it is likely to have errors. This approach works great with only 1 annotation provided, however, two problems arise when using multiple annotations. First, visible ``jumps" in quality appear when the model encounters a new annotated frame during the segmentation process, and corrects its predictions for the \textit{following} frames, but not for the \textit{previous} ones. Second, additional annotated frames were treated like ``imperfect references" - thus only having an impact on a limited part of the video, and likely to be compressed and moved to long-term memory, reducing their impact even further. 

To address these issues, we propose to add another, ``permanent" working memory module (labeled dark green in Fig.~\ref{fig:xxxmem_architecture}), implemented to store only ``ground-truth references" - i.e., annotated frames for the duration of the whole video. During inference, the annotated frames are processed separately and added to the new permanent working memory before the segmentation process starts. The ``permanent" working memory module stays unchanged throughout the whole inference, its contents are never compressed or moved to the long-term memory. During memory readout, the keys and values in the permanent memory are simply concatenated to those of the temporary memory.

This allows the model to produce smooth transitions when the target object changes between two scenes, as the model now has access to references from both (refer to Fig. \ref{fig:interpolation} from Appendix for illustration). The module also decouples the frame position from frame content, helping the model to find matches across frames regardless of their positions in the video. This results in an increased scaling of segmentation quality and efficiency, as well as fixes the ``jumping" quality issue (Section \ref{sec:results}).

\subsection{Attention-Based Frame Selection}
\label{sec:frame_selection}

Choosing the right frames to annotate leads to higher overall segmentation accuracy, which was demonstrated by previous works \cite{IVOS-W, GISRAmap, BubbleNets}. We designed an algorithm for this based on an empirical idea - to select the most \textit{diverse subset of frames} that capture the target object in different illumination conditions, pose and camera viewpoint, inspired by \cite{GISRAmap}. Given the task to select $b$ frames from a video, we assume there are $b$ different ``scenes" in it, and sample the most representative frame from each.

Given $a$ previous annotations ($a\ge 1$ since there is at least one mask provided by the user) and the predictions of the model for the rest of the frames in the video, we extract the ``key" features $k_i$ of size $(h, w)$ with our segmentation module, weighted by corresponding mask $m_i$, obtaining a region-weighted mask $r_i$. This allows the algorithm to focus on target-specific region similarity, while still having information about surrounding regions. The influence of the mask over the holistic frame features is controlled by parameter $\alpha, \alpha \in [0..1]$. With $\alpha = 0$ the algorithm ignores the annotation masks, predicting the candidates only based on the overall frame appearance, with $\alpha = 1$ it only looks at the masked regions, ignoring the rest of the pixels.
\[
    r_a = \alpha k_a \odot m_a + (1 - \alpha) k_a
\]
We then iteratively pick $b-a$ candidates with the highest dissimilarity $\mathcal{D}$, using negative pixelwise $L$-2 similarity $S$ from \cite{cheng2022xmem} with added cycle consistency. 
\[
    \mathcal{D}_{r_a, r_b} = \frac{\sum_{i, j}^{h \times w}{\max{\left(0, \left( S_{i, j}(r_a, r_b) - S_{i, j}(r_b, r_a)  \right) \right)}}}{h \cdot w}
\] 
Given frames $\mathbf{f}$ and previous annotations $\mathbf{f}$, we compute the dissimilarity across them: $\mathcal{D}_{r_ai, r_b} \forall a_i \in \mathbf{a}, b \in \mathbf{f}$. We then select $\argmax{\left( \argmin{\mathcal{D}_{r_ai, r_b}}\right) \forall a_i \in \mathbf{a}, b \in \mathbf{f}}$, the frame with the \textit{largest minimal distance} to all the existing annotations, as the next candidate. This process is repeated $b-a$ times. Due to ambiguity in pixel-to-pixel mapping, often a lot of pixels from $f_a$ map to other pixels from $f_a$, thus average self-dissimilarity is $> 0$. Cycle consistency ($S_{i, j}(r_a, r_b) - S_{i, j}(r_b, r_a)$) ensures that frames are only dissimilar if pixels in $f_i$ map to different positions in $f_j$, then from $f_j$ back to $f_i$. This guarantees that self-dissimilarity $\mathcal{D}_{r_i, r_i}=0$. Refer to the Appendix for a step-by-step explanation of the algorithm.

This allows our algorithm to demonstrate the desirable properties: it does not select candidates similar to already chosen ones, is generic and applicable to any memory-based segmentation model, and does not assume that the mask should match the visual object cues, which is violated in the case of partial segmentation, shown in Sec. \ref{sec:related_works}.

\section{Dataset and Benchmark}
 
\begin{figure}
    \centering
    \begin{subfigure}[b]{0.325\columnwidth}
        \centering 
        \includegraphics[width=\textwidth]{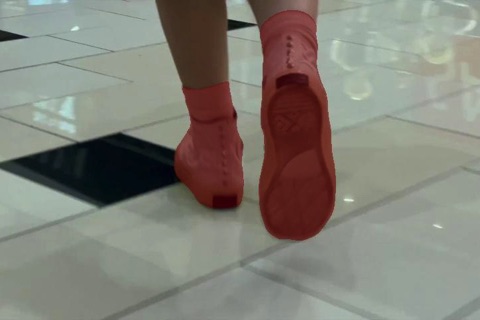} 
        \label{fig:vlogvid}
    \end{subfigure}
    \vspace{-0.5em}
    \begin{subfigure}[b]{0.325\columnwidth}  
        \centering
        \includegraphics[width=\textwidth]{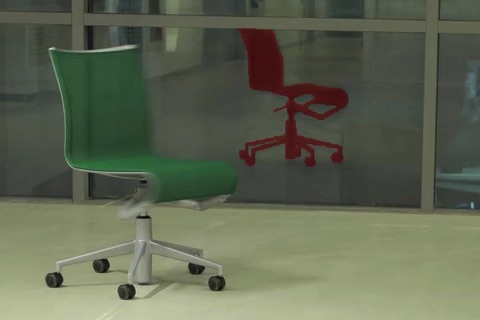} 
        \label{fig:halfface}
    \end{subfigure}
    \vspace{-0.5em}
    \begin{subfigure}[b]{0.325\columnwidth}   
        \centering 
        \includegraphics[width=\textwidth]{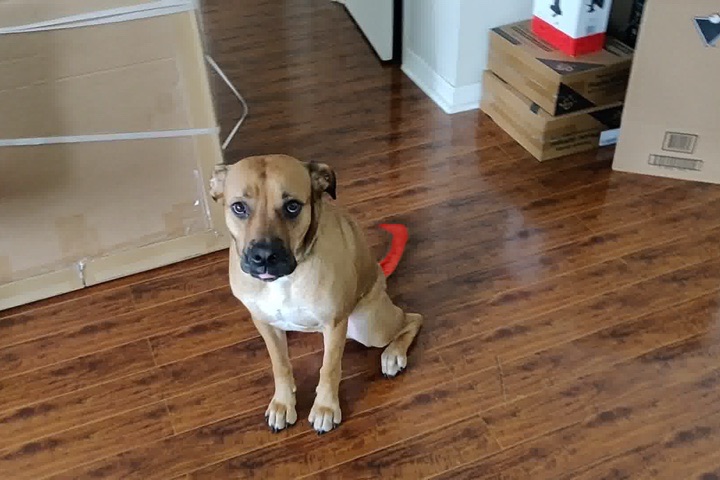} 
        \label{fig:dog_tail}
    \end{subfigure}
    \vspace{-0.5em}
        \begin{subfigure}[b]{0.325\columnwidth}   
        \centering 
        \includegraphics[width=\textwidth]{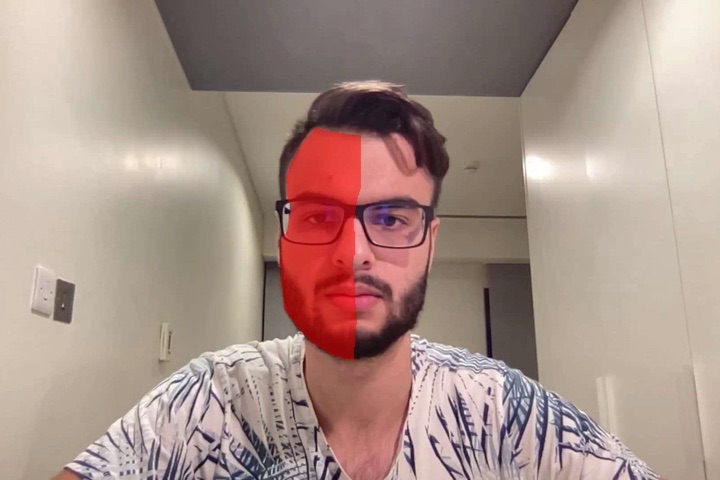} 
        \label{fig:dog_tail}
    \end{subfigure}
        \begin{subfigure}[b]{0.325\columnwidth}   
        \centering 
        \includegraphics[width=\textwidth]{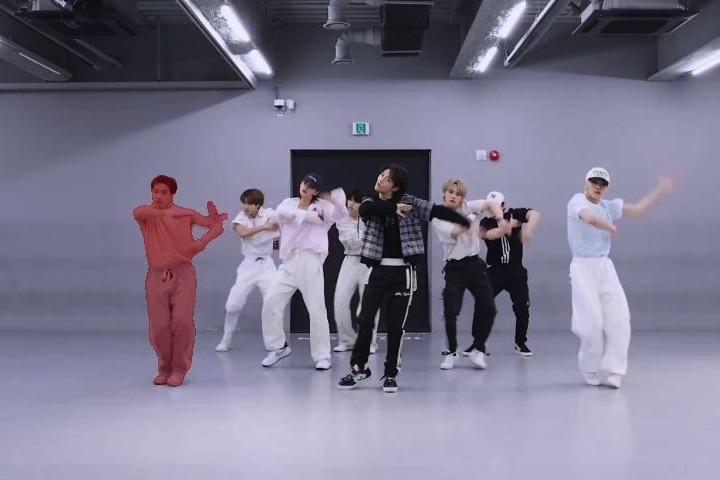} 
        \label{fig:dog_tail}
    \end{subfigure}
        \begin{subfigure}[b]{0.325\columnwidth}   
        \centering 
        \includegraphics[width=\textwidth]{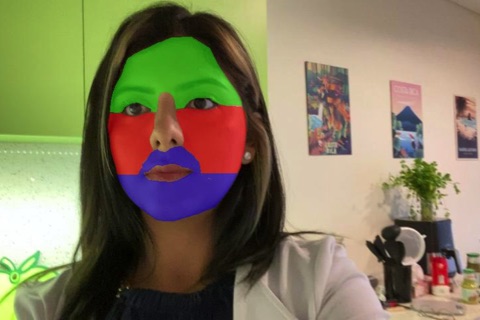} 
        \label{fig:dog_tail}
    \end{subfigure}
    \caption
    {\small Samples of our dataset (\datasetname). Best viewed zoomed. See Fig. \ref{fig:dataset_overview} in Appendix for more examples.  } 
    \label{fig:pumavosdataset}
\end{figure}

We provide a new benchmark that covers use cases for multipart partial segmentation with visually challenging situations (segments of parts of the scene with little-to-no pixel-level visual cues) called Partial and Unusual MAsk Video Object Segmentation (\datasetname). To the best of our knowledge, there are no currently available datasets like this. We contemplate scenarios from the video production industry that still conventional VOS methods struggle to tackle. We focus on partial objects such as half faces, neck, tattoos, and pimples, which are frequently retouched in film production as shown in Fig \ref{fig:pumavosdataset}. Our dataset consists of 24 clips 28.7 seconds long on average. 
To generate the annotations, we adopted a similar approach to MOSE \cite{MOSE} that used a framework with XMem \cite{cheng2022xmem} to create masks for each frame, but instead we used our method, \modelname. In MOSE the videos were annotated every 5th frame (20\% of the video), while in our case we noticed that complex scenes require 8\% to 10\% and simple scenes required 4\% to 6\% of total frames to be annotated. 


%% file: sections/experiments.tex
\section{Results}
\label{sec:results}

\begin{figure*}[h]
    \centering
    \includegraphics[width=0.8\textwidth]{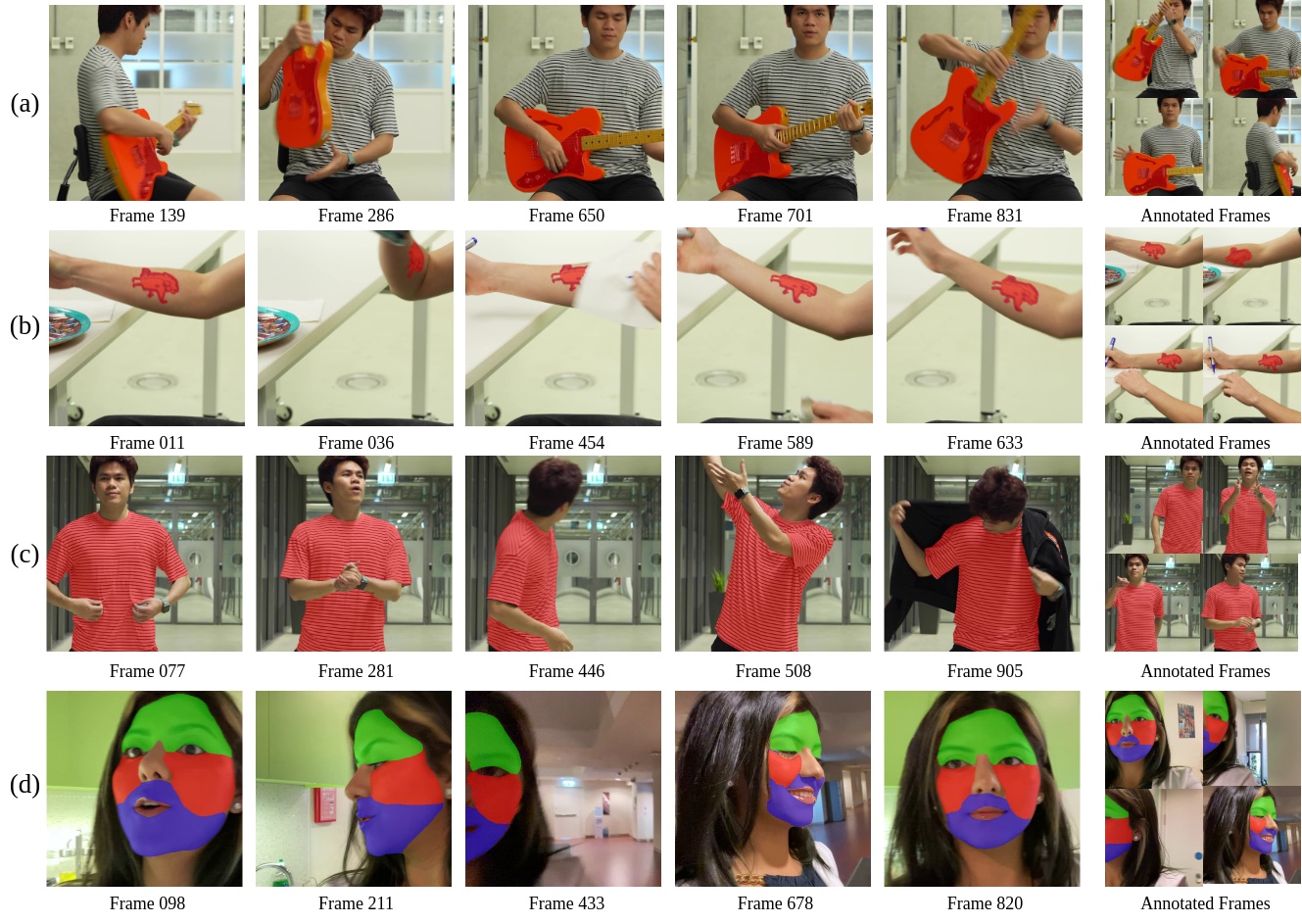}
    \caption{Results of our method \modelname{}. Here we show some use cases that can be used in the industry such as changing the front of a guitar for color purposes (row a), deformable objects such as tattoos (b) or shirts (c), multi-region with no explicit boundary (d).} 
    \label{fig:xmempp_results}
\end{figure*}

\begin{figure}[h]
    \includegraphics[width=\columnwidth]{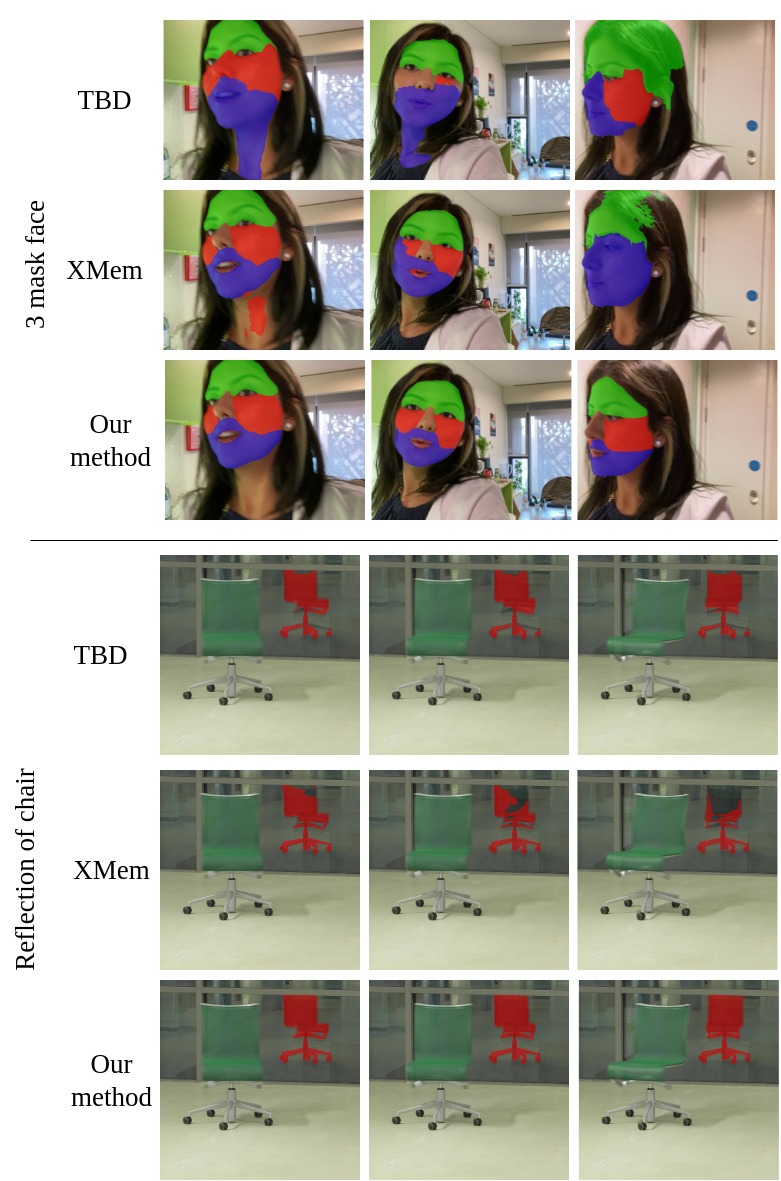}
    \caption{Comparison of our method with TBD\cite{TBD2022} and XMem\cite{cheng2022xmem} with only 6 frames provided (0.6\% of total 952 frames).}
    \label{fig:xmemppcomparison}
\end{figure}

We test our segmentation framework on a wide range of complex scenes (varying poses and deformations of human subjects, faces, rigid objects, t-shirts) and annotation tasks (object, partial, multi-class segmentation) in the presence of occlusions, lighting variations, and cropped views. In particular, we showcase rotoscoping examples that occur frequently in labor-intensive real production settings, where over 50\% of frames would need to be annotated or refined \cite{adoberotoscope2017}. Qualitative results are presented in Fig. \ref{fig:teaser}, Fig. \ref{fig:xmempp_results}, and Fig. \ref{fig:xmemppcomparison}. Our videos are recorded at 30 fps and $1920 \times 1080$ resolution, and the input is resized to $854 \times 480$ when processed by our model. We also refer to the accompanying video and supplemental material to view the results.

In row a) of Fig. \ref{fig:xmempp_results}, we show a partial segmentation result where only the front part of the guitar body is annotated, not the side or back. Only 6 out of 924 total frames were annotated (0.6\% total) in order to successfully produce reliable segmentations through a challenging sequence with frequent occlusions, pose variations, and rapid movement.
Rows 6 in Fig. \ref{fig:xmemppcomparison} depicts an example where the goal is to composite the reflection of an object (e.g., a chair) into a different scene, which is a common use case in visual effects production. Our chair rotates throughout the video, which changes its projected outline significantly, but a consistent segmentation can be performed by only annotating 6 frames out of 411 total frames (1.5\%). This use-case is especially challenging since the scene contains a larger, more prominent object with the exact same appearance and movement. A challenging multi-class segmentation example for a long and diverse sequence is illustrated in row 3 of Fig. \ref{fig:xmemppcomparison}, where the subject's face is segmented into 3 different regions, one of which consists of two separate parts. All regions are segmented simultaneously, which prevents them from overlapping with each other which can happen when each region is segmented independently. The additional challenge in this scenario is that the regions are often not separated by prominent visual cues such as boundaries, corners, edges, etc. Highly consistent results can be extracted even in the presence of extreme lighting variations, moving background and head poses, where again only 6 out of 951 (0.6\%) total frames have been annotated. 

\paragraph{Evaluation.}
We evaluate our framework on a diverse set of complex videos from MOSE \cite{MOSE} dataset, as well as on long videos provided by LVOS \cite{LVOS}. We use the training subset of MOSE (since it does not provide annotation for the test subset), consisting of 1507 videos, and a validation subset of LVOS consisting of 50 videos\footnotemark. We compare the performance of multiple SSVOS models when given $1$, $5$, and $10$ uniformly-sampled annotated frames. For comparison, we picked existing works in SSVOS and IVOS, that support the usage of multiple annotation frames by design as well as support dense annotations, since MOSE and LVOS do not provide scribbles information.

The performance of \modelname{} with only one annotation given is equivalent to XMem. We see that on both datasets \modelname{} demonstrates noticeably better performance scaling in terms of $\mathcal{J}$ and $\mathcal{F}$ metrics, for 5 and 10 annotated frames provided at input. Moreover, it can be seen that \modelname{} achieves comparable or higher segmentation quality with fewer annotations provided (5) than the competition (10), thus making it on average $2\times$. Furthermore, we evaluate our model and XMem on a subset of \datasetname{} dataset and observe that on some videos the efficiency of \modelname{} is up to $5 \times$ higher. 

\begin{figure}[h]
    \centering
    \includegraphics[width=0.995\columnwidth]{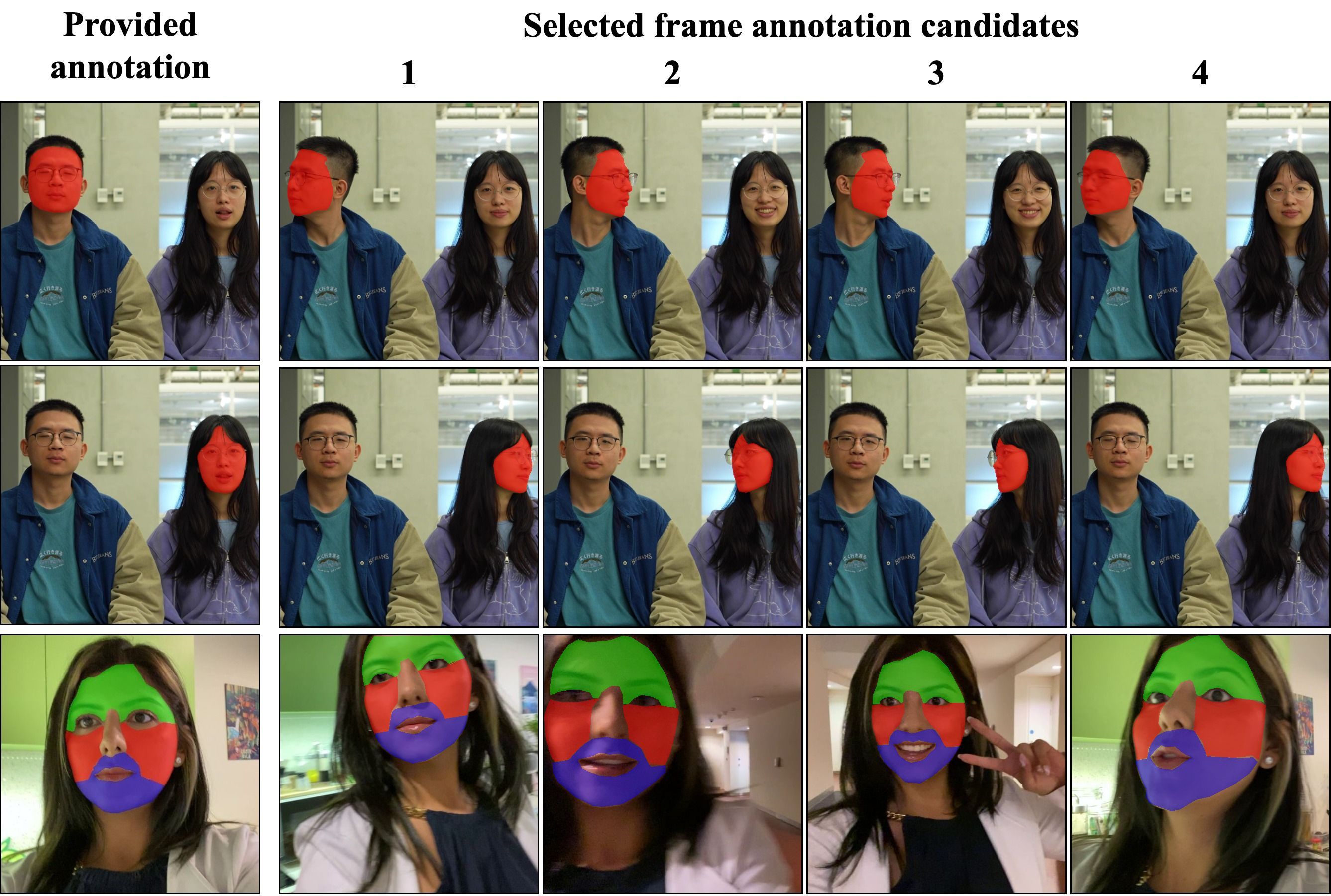}
    \caption{Demonstrated behavior of our target-aware frame selection algorithm. Rows 1 and 2 depict the same video sequence, but with different target annotation. Numbers indicate the importance ranking produced by the algorithm.}
    \label{fig:frame_selector}
\end{figure}

\begin{table}[t]
\setlength\tabcolsep{1.75pt}
\renewcommand{\arraystretch}{1.25} 
\centering
\small
\begin{tabular}{c|cc|cc|cc|cc}
\cline{2-9}
\multicolumn{1}{l}{}            & \multicolumn{6}{c|}{Number of annotations provided}   & \multicolumn{2}{c}{\multirow{2}{*}{$|\mathcal{D}_{1 \rightarrow 10}|$}}   \\  \cline{2-7}
\multicolumn{1}{l}{}            & \multicolumn{2}{c|}{1 frame}   & \multicolumn{2}{c|}{5 frames}  & \multicolumn{2}{c|}{10 frames} & & \\  \hline

\textbf{Method}                 & $\mathcal{J}$             &$ \mathcal{F}$             & $\mathcal{J}$ & $\mathcal{F}$ & $\mathcal{J}$ & $\mathcal{F}$  & $\mathcal{J}$ & $\mathcal{F}$ \\  \hline

$\text{TBD}_{\textit{DAVIS}}$   & 42.72                     & \textbf{53.38}            & 52.17         & 63.64         & 60.04         & 71.65          & \good{+17.3}         & \good{+18.3}         \\
$\text{TBD}_{\textit{YT}}$      & 41.15                     & 50.79                     & 57.19         & 67.83         & 65.07         & 75.72          & \good{+23.9}         & \good{+24.9}         \\
XMem                            & \multirow{2}{*}{\textbf{44.06}} & \multirow{2}{*}{52.33}& 56.30       & 66.05         & 62.62         & 73.18          & \good{+18.5}         & \good{+20.9}        \\
\modelname                      &                           &                           & \textbf{67.36}& \textbf{78.11}& \textbf{75.74}& \textbf{86.35} & \textbf{\good{+31.7}}& \textbf{\good{+34.0}} \\ \hline
\end{tabular}

\caption{Quantitative results on LVOS \cite{LVOS} validation dataset. $\mathcal{J}$ and $\mathcal{F}$ mean Jaccard index and boundary F-score correspondingly, as defined in \cite{DAVIS}. $\text{TBD}_{\textit{DAVIS}}$ and $\text{TBD}_{\textit{YT}}$ stand for TBD \cite{TBD2022} model trained on DAVIS \cite{DAVIS} and YouTube-VOS \cite{YOUTUBE-VOS} datasets accordingly. \textit{At $k=5$ annotation frames \modelname{} achieves higher quality ($\mathcal{J}$ and $\mathcal{F}$) then all other models at $k=10$ frames.} $|\mathcal{D}_{1 \rightarrow 10}|$ denotes the increase in segmentation quality from $1$ to $10$ annotated frames. }
\label{tab:LVOS_scaling}
\end{table}

\footnotetext{One of the videos provided does not have any target objects on frame \#0, which an unsupported use-case for some of the models used, so only 49 out of 50 videos were included in the evaluation. }

\begin{table}[h]
\setlength\tabcolsep{1.75pt}
\renewcommand{\arraystretch}{1.25} 
\centering
\small
\begin{tabular}{c|cc|cc|cc|cc}
\cline{2-9}
\multicolumn{1}{l}{}            & \multicolumn{6}{c|}{Number of annotations provided}   & \multicolumn{2}{c}{\multirow{2}{*}{$|\mathcal{D}_{1 \rightarrow 10}|$}}   \\  \cline{2-7}
\multicolumn{1}{l}{}            & \multicolumn{2}{c|}{1 frame}   & \multicolumn{2}{c|}{5 frames}  & \multicolumn{2}{c|}{10 frames} & & \\  \hline

\textbf{Method}                 & $\mathcal{J}$             &$ \mathcal{F}$             & $\mathcal{J}$ & $\mathcal{F}$ & $\mathcal{J}$ & $\mathcal{F}$  & $\mathcal{J}$ & $\mathcal{F}$ \\  \hline

$\text{TBD}_{\textit{DAVIS}}$   & 43.34                     & 49.70            & 56.57         & 63.00         & 63.15         & 69.15          & \good{+19.8}         & \good{+19.5}          \\
$\text{TBD}_{\textit{YT}}$      & 48.28                     & 54.05                     & 62.71         & 68.94         & 68.24         & 74.23          & \good{+20.0}         & \good{+20.2}          \\
STCN                            & 54.51 &                   60.69                       & 58.73         & 65.86         & 62.78         & 70.11          & \good{+8.3}         & \good{+9.4}          \\
XMem                            & \multirow{2}{*}{\textbf{57.21}} & \multirow{2}{*}{\textbf{63.98}}& 67.95 & 76.41      & 77.78         & 85.26          & \good{+20.6}         & \good{+21.3}          \\
\modelname                      &                           &                           & \textbf{77.11}& \textbf{84.56}& \textbf{82.87}& \textbf{90.20} & \textbf{\good{+27.7}}& \textbf{\good{+26.5}} \\ \hline
\end{tabular}
\caption{Quantitative results on MOSE \cite{MOSE} training dataset. $\mathcal{J}$ and $\mathcal{F}$ mean Jaccard index and boundary F-score correspondingly, as defined in \cite{DAVIS}. $\text{TBD}_{\textit{DAVIS}}$ and $\text{TBD}_{\textit{YT}}$ stand for TBD \cite{TBD2022} model trained on DAVIS \cite{DAVIS} and YouTube-VOS \cite{YOUTUBE-VOS} datasets accordingly. Since the training subset of MOSE dataset includes some very short videos, we only considered videos with $>= 50$ frames each for comparison. Results from a total of $722$ videos are presented. $|\mathcal{D}_{1 \rightarrow 10}|$ denotes the increase in segmentation quality from $1$ to $10$ annotated frames.}
\label{tab:MOSE_scaling}
\end{table}

\begin{table}[h]
\setlength\tabcolsep{1.75pt}
\renewcommand{\arraystretch}{1.25} 
\centering
\small
\begin{tabular}{c|cc|cc|cc|c}
\cline{2-8}
\multicolumn{1}{l}{}            & \multicolumn{2}{c|}{XMem}   & \multicolumn{2}{c|}{\modelname{}}  & \multicolumn{2}{c|}{\# Annotations} & \multirow{2}{*}{$\mathbb{E}$} \\  \cline{2-7}

\textbf{Sequence}                 & $\mathcal{J}$             &$ \mathcal{F}$             & $\mathcal{J}$ & $\mathcal{F}$ & XMem & \modelname{}  \\  \hline

Vlog 3-part     & 85.43          & 89.70  & \textbf{85.45} & \textbf{89.89} & 45             & 10         & $4.5 \times$             \\
Lips            & \textbf{87.24} & \textbf{94.79} & 86.93          & 94.43          & 45         & 10          & $4.5 \times$                  \\
Half face       & 93.26          & 98.13          & \textbf{93.31} & \textbf{98.35} & 50         & 10          & $5 \times$             \\  \hline
\end{tabular}

\caption{Quantitative results on a subset of \datasetname{} dataset. $\mathcal{J}$ and $\mathcal{F}$ mean Jaccard index and boundary F-score correspondingly, as defined in \cite{DAVIS}. $\mathbb{E}$ is the frame usage efficiency of \modelname{} compared to XMem.}
\label{tab:pumavos_scaling}
\end{table}
The behavior of our annotation candidate selection module is demonstrated in Fig. \ref{fig:frame_selector}. Often there is more than one possible target in the video, so selecting frames for the right one is important. A video is presented in rows 1 and 2 of Fig. \ref{fig:frame_selector}, where two people change their head pose, but one at a time. Our algorithm successfully adapts the recommended frames based on which person is being segmented. Rows 3 and 4 depict a more complicated video sequence, where the target object has extreme lighting variations, rapid movement, disappearances, pose and expression variations (whenever applicable). The algorithm selects frames that capture these varieties without repeating or selecting frames with no target object. Our experiments indicate that the selected annotation candidates are very similar to those chosen by expert users, which makes it practical in real-life scenarios where an end user is likely to work with multiple videos at the same time, thousands of frames long, and rewatching them for multiple rounds to select the frames manually is often infeasible.

\paragraph{Comparison.}
We compare our results with 3 SOTA methods in Table \ref{tab:MOSE_scaling}, and with 2 in Table \ref{tab:LVOS_scaling}, and Fig. \ref{fig:xmemppcomparison}. We demonstrate that our model produces smooth and temporally continuous segmentation in rows 3-6 of Fig. \ref{fig:xmemppcomparison}, where both other methods produce incorrect masks missing a part of the target object. In rows 1-3 of Fig. \ref{fig:xmemppcomparison} our method successfully segments challenging multi-part regions of the face, that are mostly not visually aligned with the low-level image cues, and the masks produced by TBD and XMem are ``bleeding" into the neighbouring regions, as well as have sharp, ``torn"-looking edges. In Tables \ref{tab:MOSE_scaling} and \ref{tab:LVOS_scaling} we demonstrate that our method results in higher segmentation quality given the same frame annotations, and is at least $2 \times$ as efficient in the number of annotations necessary on generic videos. We additionally show in Table \ref{tab:pumavos_scaling} that in more challenging practical sequences \modelname{} can be up to $5 \times$ more efficient.

\paragraph{Limitations.}

The segmentation quality of our method sometimes suffers on blurry frames with quickly moving objects, and the similarity measure for the annotation candidate selection is not well-defined on such data either. With multiple similar/identical objects in the frame, the method can sometimes switch to the wrong target if they occlude each other. Use cases of extreme deformation (clothing) and high-detail objects (hair) remain an active challenge. Visual illustrations are provided in the Appendix.

\paragraph{Performance.}

In the original XMem, given $n$ frames, the processing time is bound by $O \left( \frac{n^2k}{z} + \frac{n}{q} \right)$, where $k$ is the maximum size the working memory (typically $k=100$), $q$ is the memory insertion frequency (typically $q=5$), and $z$ is a compression rate for long-term memory ($z > 600$) \cite{cheng2022xmem}. 

Given $m$ annotated frames, \modelname{} loads them into the permanent memory (static $+m$ factor), with the working memory size $=k+m$, thus processing time is $O\left( \frac{n^2(k+m)}{z} + \frac{n}{q} + m \right)$. In practice $m$ is likely to be small, $m \leq 20$, thus $m \leq 0.2 k$, having a slowdown of $<1.2\times$ on memory readout, and an even smaller effect on the overall segmentation process. On RTX-3090 GPU with a $500$-frame video and $5$ annotations provided, at $854 \times 480$ resolution, \modelname{} yields 32 FPS ({35} FPS excluding loading the frames into permanent memory), and XMem yields {39} FPS. Total memory usage only increases by a static factor of $+m$, as we store $m$ additional annotations.

%% file: sections/conclusion.tex
\section{Discussion}

We introduced a highly robust semi-supervised and interactive video segmentation framework, \modelname, with automatic next best frame prediction for user annotation. We have shown that by introducing a permanent memory module to XMem~\cite{cheng2022xmem}, efficient usage of multiple annotated frames is possible, for segmenting a particular object or region, even with drastic changes in the appearance of the object. Our approach achieves better segmentation results than current SOTA VOS methods with significantly fewer frame annotations (in our experiments, up to 5$\times$ fewer annotations in highly challenging cases). Our approach further demonstrates the ability to reliably segment partial regions of an object (e.g., the left half of a face) with only a few frame annotations, which is a notoriously difficult task for any existing segmentation methods. As highlighted in our accompanying video, even for highly challenging and long scenes, our masks are temporally smooth without the need for additional post-processing. Hence, our method is suitable for production use cases, such as rotoscoping, where accurate region segmentation and minimal user input is needed.

Our proposed solution is also suitable for non-expert users, as it suggests the next best frame for the user to annotate using an effective yet simple attention-based algorithm. Our experiments indicate that the predicted frames are often very similar to those chosen by expert users, which is always superior to randomly chosen ones. We also show that our framework can be conveniently used to collect and annotate a new dataset, \datasetname{}, covering challenging practical segmentation use-cases, such as partial segmentations, multi-object-part segmentation, complex lighting conditions, which cannot be found in existing datasets. 

\paragraph{Future Work.}

While our framework significantly improves the current SOTA within the context of IVOS, we believe that further reduction in frame annotations and complex shape segmentations is possible. In particular, we plan to investigate methods that incorporate dense scene correspondences and on-the-fly generative data augmentation of the segmented regions, which can even be used to improve the robustness of the frame prediction further. 

%% file: sections/appendix.tex
\appendix
\section{\datasetname{} Dataset}
\datasetname{} is a dataset that covers visually challenging segmentation scenarios, including, but not limited to high appearance variability due to changes in lighting, viewing angles, deformation, and scale changes; partial segmentation (where only a part of the object is being segmented), often with limited to no low-level image cues (e.g. half of a person's face) and occlusion. It consists of 24 videos, from 13.5 to 60 seconds long, 28.7 on average, with 480p resolution with different aspect ratios (both vertical and horizontal). 

\begin{figure*}[!htb]
    \centering
    \makebox[\textwidth][c]{\includegraphics[width=1\linewidth]{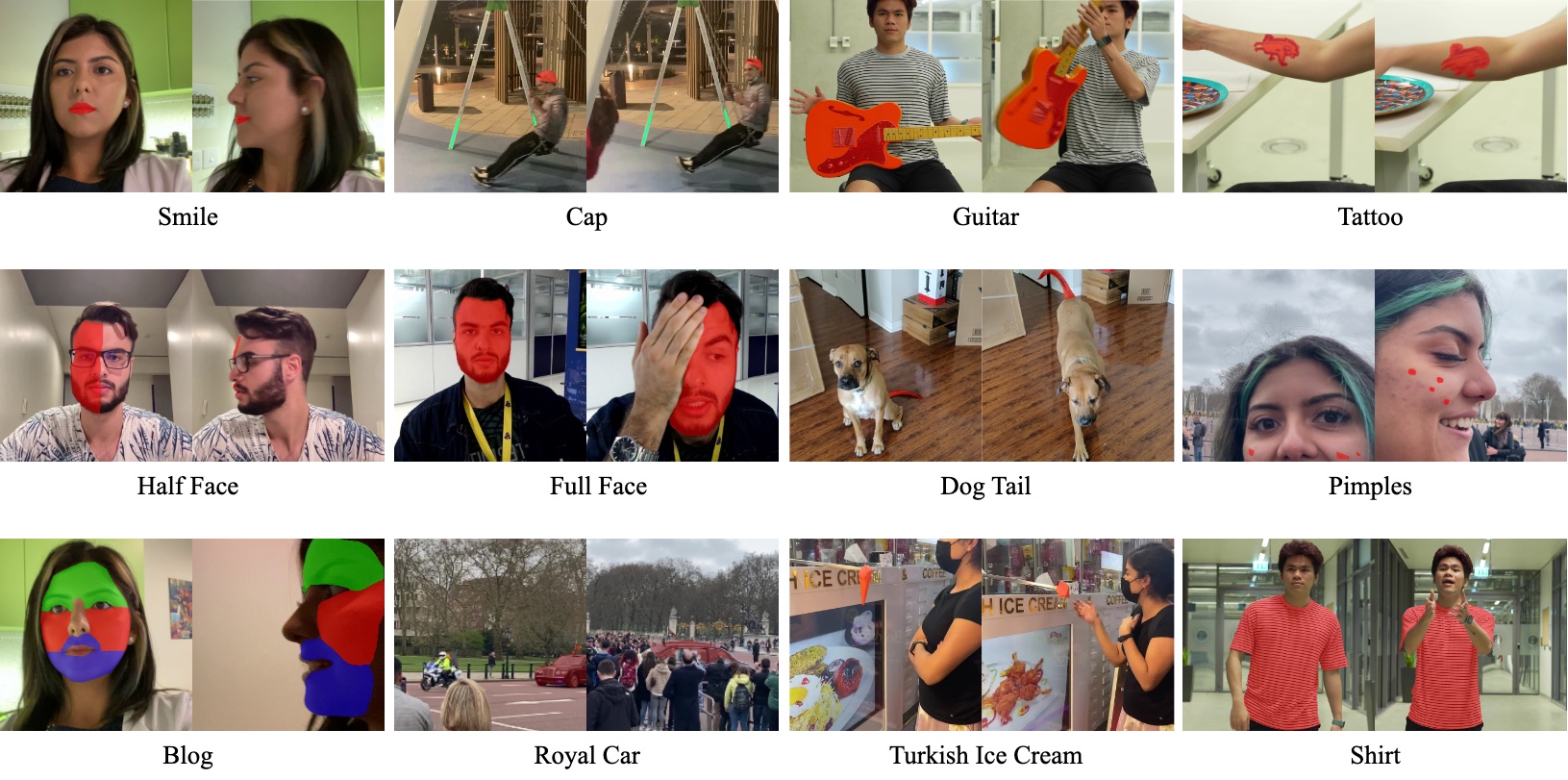}}
    \caption{\datasetname{} dataset overview}
    \label{fig:dataset_overview}
\end{figure*}

Examples of the videos from the dataset are illustrated in Fig. \ref{fig:dataset_overview}. Sequences ``Half Face", ``Guitar" and ``Dog Tail" depict challenging partial segmentation use-case where only the left part of a person's face, the dog's tail or just the body of the guitar body is segmented, without following any image cues such as edges or corners. ``Royal car" and ``Full Face" contain changes in scale and heavy occlusion throughout the sequence, while ``Pimples" and ``Turkish Ice Cream" both contain very small objects that also move a lot throughout the scene. ``Pimples" and ``Cap" are also examples of ambiguity - there are multiple objects with similar appearances present in the video, but only some of them are supposed to be segmented. ``Vlog" is one of the most challenging sequences in the dataset, as it not only contains partial segmentation with arbitrarily-defined boundaries (irrespective of the low-level image cues) but also has multiple target regions to segment, which are located on the same physical object, as well as having a lot of variability in lighting, static and dynamic background scenes, and frequent deformations, which are also present in ``Shirt" and ``Tattoo" sequences. 

\textit{\datasetname{} and the source code of \modelname{} and the annotation algorithm are going to be released to the public after the paper is accepted. We also provide pseudocode for our frame selection algorithm.}

\section{Frame Selection User Study}
A user study was performed to analyze the effectiveness of the automatic frame selection module. A total of seven participants were selected - two experienced and two novice users. The users were given three videos each and asked to select 5 most representative frames (besides frame \#0) that capture the variation in the appearance of the target object. The videos chosen from \datasetname{} are: sequences ``Half Face" (shortened), ``Turkish Ice Cream", and ``Vlog". Videos last from 22s (673 frames) to 32s (960 frames). 

The experienced users were explained how \modelname{} works internally in detail, as well as shown its predictions on a few sample videos, while novice users treated it as a blackbox. The participants were shown the full video and the annotation of the target object(s) for the first frame and were free to rewatch it as many times as they deemed necessary, then asked to pick the frames. They were told that the relative position of the frame does not matter, only its content. We recorded the time it took the users to select the frames (excluding the initial video viewing time) and compared it to the running time of the frame selection algorithm. We then took the chosen annotations and performed a quantitative comparison of speed and final segmentation accuracy using \modelname{}, presented in Table \ref{tab:user_study}. 

\begin{table}[ht]
\centering
\begin{tabular}{l|ccc}
\hline
\textbf{Group/Method} & \textbf{IoU} & \textbf{F-score} & Average time, s \\
\hline
Algorithm & 0.777 & 0.828 & \textbf{1.7} \\
Experts & \textbf{0.805} & \textbf{0.855} & 47.1 \\
Non-experts & 0.779 & 0.825 & 54.8 \\
\hline
\end{tabular}
\caption{Comparison of performance metrics across expert and non-expert with the frame annotation candidate selection algorithm.}
\label{tab:user_study}
\end{table}

Our algorithm is $27 \times$ faster than the expert users, and $32 \times$ faster than non-expert ones, while providing comparable performance to non-expert users' results. This makes it a practical tool for large-scale environments where it is infeasible to have a lot of trained experts to work on videos, while also having a practical application for expert users, who can use the algorithm to very quickly obtain a lot of potentially valuable annotation candidates and then select the best ones with their expertise, saving a lot of time in the process.
\section{Additional Results}
\begin{figure*}[!htb]
    \centering
    \makebox[\textwidth][c]{\includegraphics[width=1\linewidth]{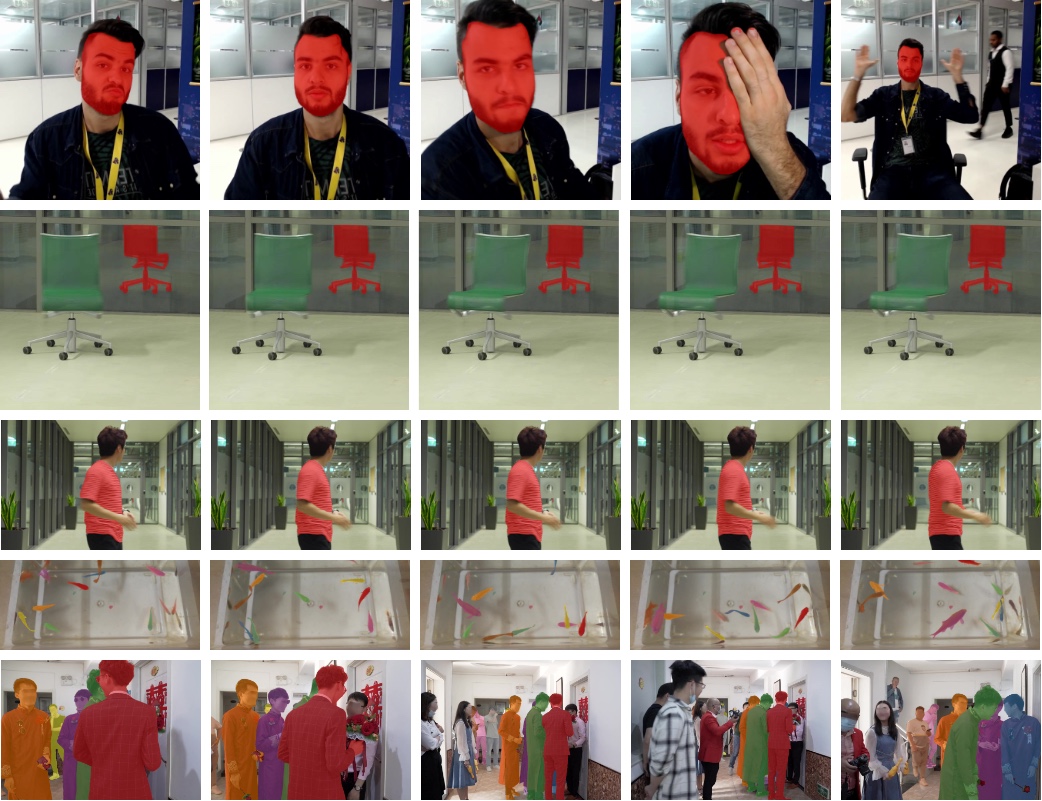}}
    \caption{Results of XMem++}
    \label{fig:good_results}
\end{figure*}

Please refer to the accompanying video to see all the video results and method comparisons. We highlight a number of highly complex scenes, where existing methods are challenged with varying scales, appearance, occlusions, and ambiguous objects. For example, in Fig. \ref{fig:good_results} row 1 the subject's face is segmented (without ears or hair), while they are going through a variety of poses, rotate around, occlude the target region, and move both closer to and farther from the camera. 

Our method successfully segments the target across multiple scales and poses, resulting in a smooth, temporally coherent, and accurate segmentation. Rows 4 and 5 depict scenes with similar challenges - a multitude of objects present, that have a similar appearance, frequently occlude each other and move, both around the scene and with their individual body parts. In both cases, our method successfully segments all of the targets, without confusing them, ``bleeding" the mask into neighboring objects, or merging multiple targets into one. 

Moreover, in the last picture of row 5, Fig. \ref{fig:good_results} it can be observed that the flower the person marked with the blue mask is holding was correctly not segmented, since in the provided annotations, it is not included, as not being a part of the target object. This illustrates that \modelname{} can work correctly with a large number of targets in the scene while preserving a high level of detail about their appearance.

\begin{figure}[!h]
    \centering
    \includegraphics[width=1\linewidth]{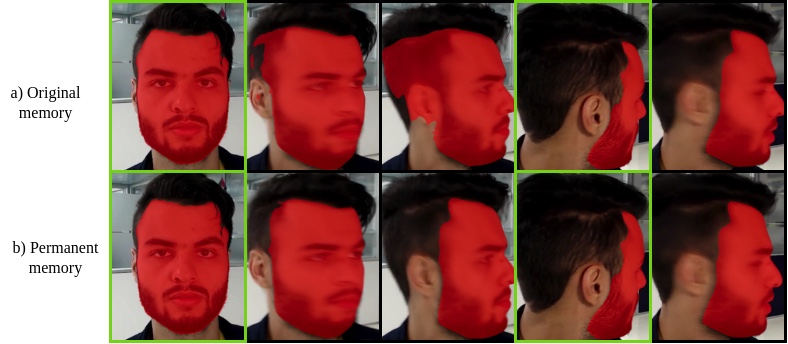}
    \caption{Illustration of smooth interpolation between different object's appearance that the permanent memory module in \modelname{} provides. The green frames are the ground truth annotations given to the model. It is noticeable that \modelname{} (row b) smoothly interpolates the mask across the change in the face's orientation, but the original XMem (row a) only fixes its predictions after processing the second ground truth annotation, resulting in a sharp "jump" in visual quality.}
    \label{fig:interpolation}
\end{figure}

\begin{figure*}[!htb]
    \centering
    \makebox[\textwidth][c]{\includegraphics[width=1\linewidth]{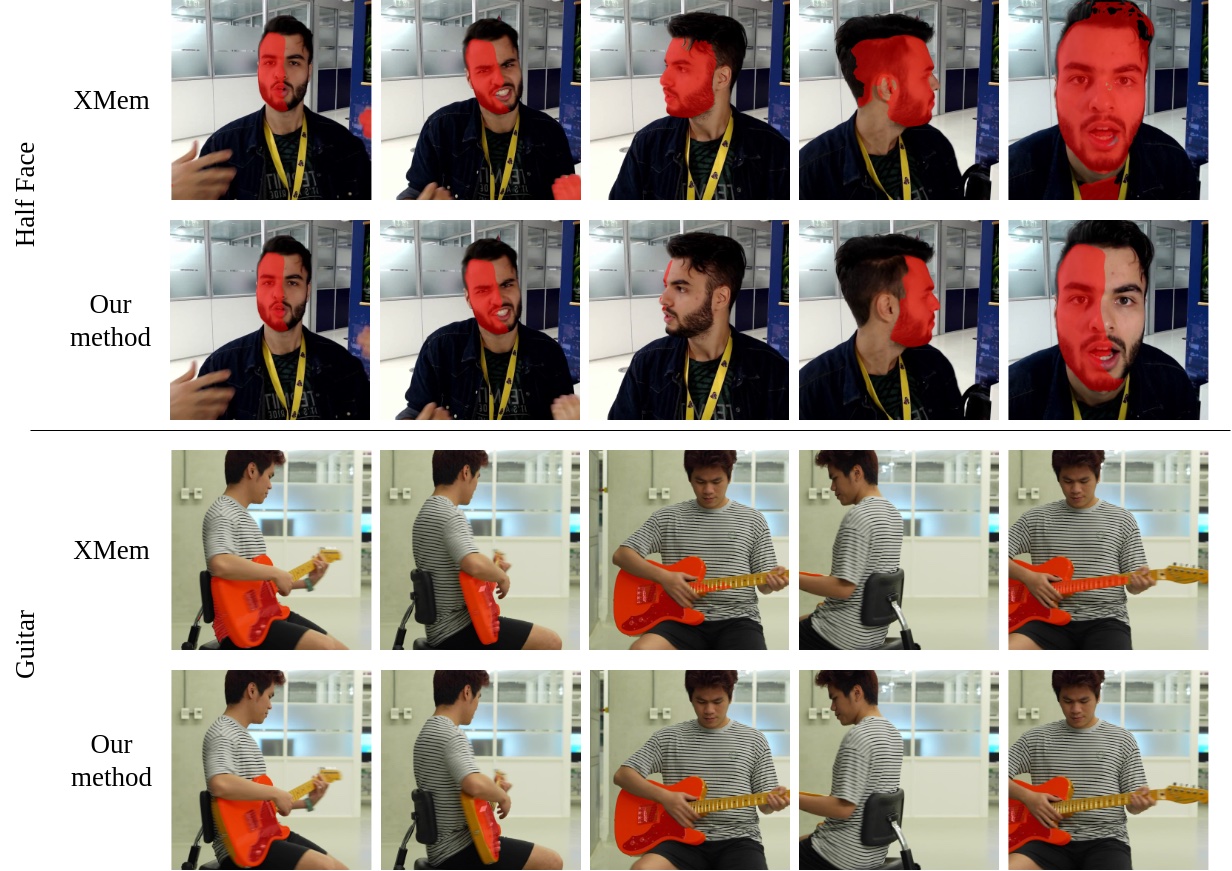}}
    \caption{Comparison models}
    \label{fig:comparison}
\end{figure*}

\paragraph{Comparisons.}
We provide additional comparisons between our method and the current SOTA interactive segmentation model XMem \cite{cheng2022xmem}. XMem is a resource-efficient and fast memory-based segmentation method introduced in 2022 by Ho Kei Cheng and Alexander G. Schwing. Additional comparison results are provided in Fig. \ref{fig:comparison} and Fig. \ref{fig:interpolation}. For each video 6 frames were selected for annotation by uniformly sampling the video (refer to Eq. \ref{eq:uniform_baseline}), starting from frame 0. Row 1 shows the same video as in ``Full Face" sequence from \datasetname{}, but now with only half of the face being segmented, thus providing the same challenges as discussed earlier, but with even more difficult segmentation. Row 2 is a ``guitar" example from \datasetname{}, where only the frontal part of the guitar's body is the target. In both cases we see \modelname{} resulting in a noticeably better segmentation, in particular with the ``Half face" sequence, where it manages to produce the correct segmentation mask throughout the extreme variations in pose, expression, and scale, while XMem often ``bleeds" the mask into neighboring regions, sticking more to the visual cues of the object. The results for the ``Guitar" sequence demonstrate a similar outcome - \modelname{} correctly segmenting the front of the guitar, but not the sides, while XMem segments the whole object, again overfitting to visual cues instead of correct segmentation boundaries.

\begin{figure*}[!htb]
    \centering
    \makebox[\textwidth][c]{\includegraphics[width=0.6\linewidth]{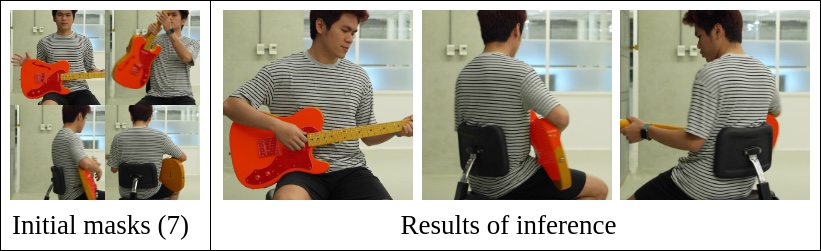}}
    \caption{Limitation of \modelname{}: Partial failure (the back of the guitar is segmented in some frames)}
    \label{fig:limitations}
\end{figure*}

%
%
\paragraph{Limitations.}
Some examples are challenging even for our method. For example, while some motion blur is fine, with extreme motion blur it can't, which is a common challenge for all existing methods. Furthermore, memory-based models generally struggle when provided ``negative masks" - empty annotations where the target object is not present, but the model has a false-positive segmentation prediction, illustrated in Fig. \ref{fig:limitations}

\begin{figure}[!htb]
    \centering
    \includegraphics[width=0.75\linewidth]{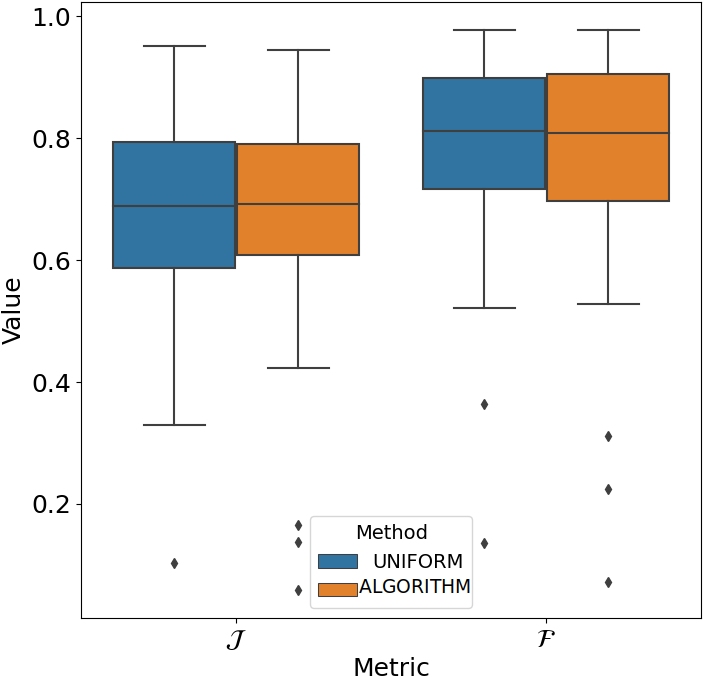}
    \caption{Comparison of segmentation quality with frames chosen by uniform baseline and our candidate selection algorithm on LVOS validation dataset.}
    \label{fig:lvos_candidates}
\end{figure}

Our frame selection algorithm does not work well when there is too much dynamics in the scene, with a lot of objects moving chaotically at the same time. Equivalently, if there is too little movement, no clear scene boundaries, or very little variation in the target object's appearance. In both of these cases, the importance of selecting the right candidates for annotations is significantly reduced, as most of the frames result in a similar accuracy improvement. In this case, our algorithm wouldn't necessarily perform better than randomly/uniformly selected frames. To prove this, we evaluate our frame selection algorithm with \modelname{}, and a uniform baseline, calculated by the formula in the Eq. \ref{eq:uniform_baseline} on LVOS dataset \cite{LVOS}, in which the videos typically have one of the aforementioned traits, and we show that the performance of the uniform baseline and our algorithm is very similar (Fig. \ref{fig:lvos_candidates}). 

\begin{equation}
\label{eq:uniform_baseline}
    F_A = \lfloor linspace(0, N - 1, k) \rfloor
\end{equation}

Given a video with $N$ total frames, we select $k$ candidates for both uniform and our frame annotation candidate selection algorithm. We then run inference on LVOS validation set with 49 videos, described in the Results section in the main paper, and illustrate the distribution of both F-score ($\mathcal{F}$) and Intersection-over-Union ($\mathcal{J}$) metrics in Fig. \ref{fig:lvos_candidates}.

\section{Additional Evaluation}
\begin{figure}[!htb]
    \centering
    \includegraphics[width=1\linewidth]{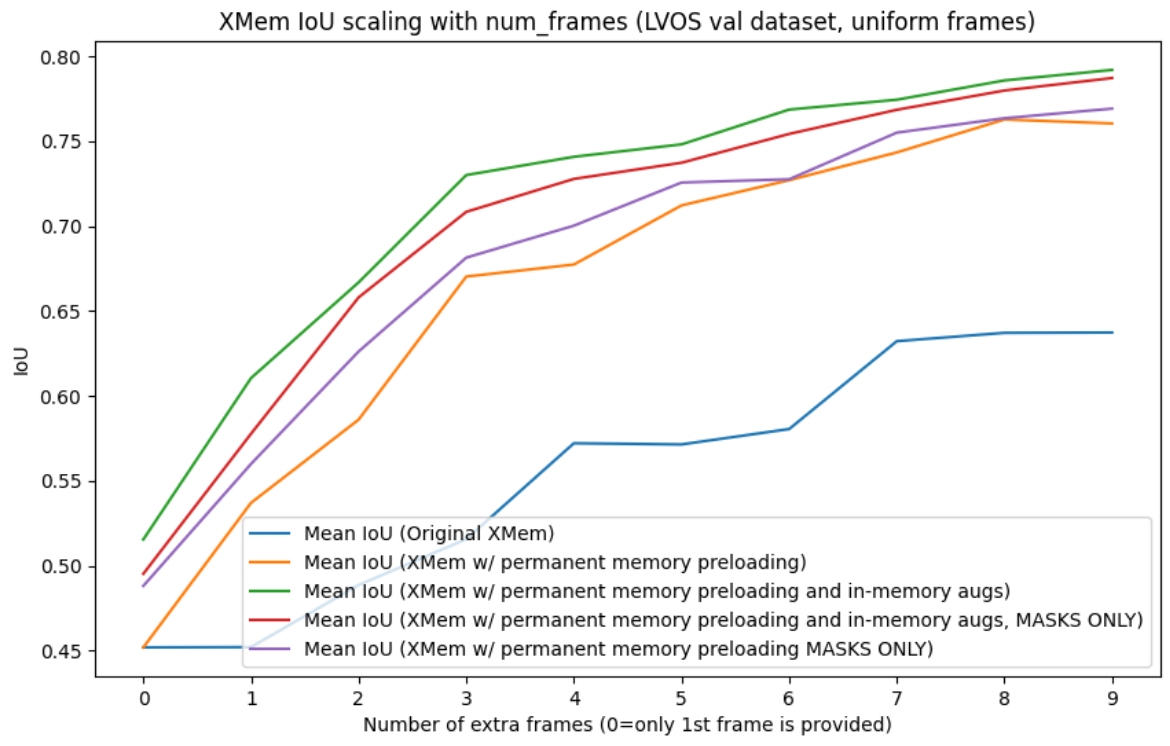}
    \caption{Demonstration of superior segmentation quality scaling of \modelname{} compared to original XMem with the number of annotated frames available. Blue line is the original XMem model, orange is \modelname{} as used in all other evaluations and comparison. Purple line shows a modification of \modelname{} with disabled temporary memory, and green and red indicate the usage of in-memory augmentations, with and without temporary memory correspondingly.}
    \label{fig:scaling}
\end{figure}

\begin{figure*}[!htb]
    \centering
    \makebox[\textwidth][c]{\includegraphics[width=1\linewidth]{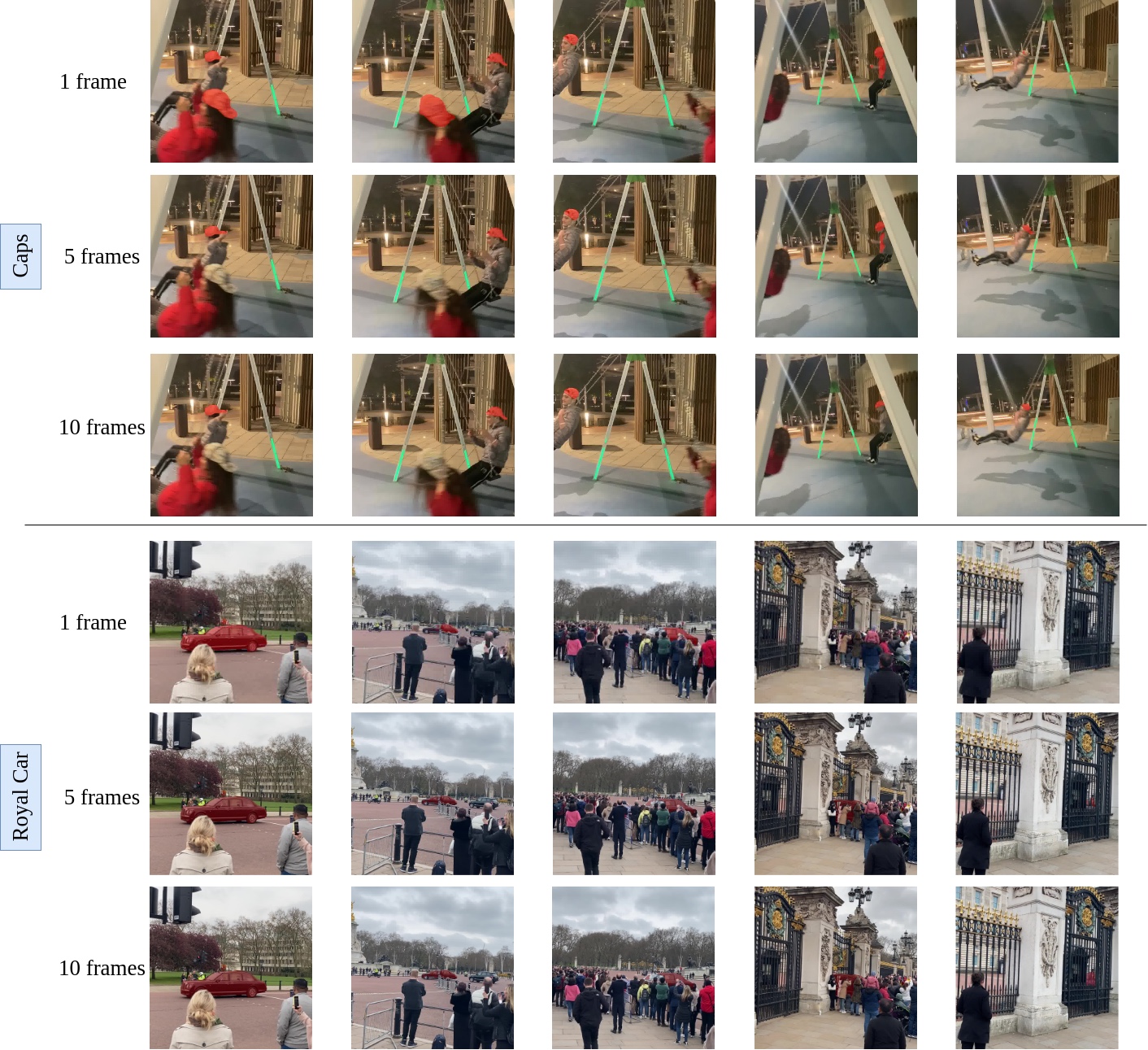}}
    \caption{Visual comparison of the segmentation results by \modelname{} with 1, 5, and 10 uniformly-sampled annotation candidates provided.}
    \label{fig:1_5_10}
\end{figure*}

We analyze the performance increase of XMem and \modelname{} with the number of annotations available, by providing both qualitative (Fig. \ref{fig:1_5_10}) and quantitative results (Fig. \ref{fig:scaling}). In Rows 1-3 of the ``Caps" sequence in Fig. \ref{fig:1_5_10}, we see that providing just 5 frames is enough to completely resolve the ambiguity problem when segmenting one of the two identical caps in the frame. Providing 10 annotated frames further improves segmentation quality in challenging scenes, such as in Columns 4-5, where there is a lot of motion blur on the target object, as well as lighting variation. \modelname{} demonstrates similar results for ``Royal car" sequence where just 5 provided annotated frames drastically improve the quality in the scenes with extreme occlusion, where the car is hardly visible behind lots of people (Columns 2-4).

We furthermore evaluate our method's quality scaling performance on LVOS validation dataset, from 1 to 10 annotated frames provided, illustrated in Fig. \ref{fig:scaling}. As expected, given only 1 frame both models yield equivalent results, however \modelname{} (drawn with orange line) demonstrates significantly higher scalability potential and efficiency starting at 2 annotated frames and keeps the advantage throughout the whole comparison, up to \textbf+{13\%} difference at 10 frames ($0.63$ XMem vs $0.76$ \modelname{})

\paragraph{In-Memory Augmentations.}
We address a possible use-case where annotations could be very sparsely available or too expensive to produce, by exploring in-memory augmentations for provided annotated frames. Through rigorous testing, we select the \textbf{11} best augmentations that, when combined, lead to the highest possible segmentation quality improvement for \modelname{}, shown in Fig. \ref{tab:augs}. When processing and adding provided frames and their annotations to the permanent memory, each of the augmentations is applied to every frame (and their corresponding mask, where necessary) and stored in the permanent memory as well. This can be a practical way of increasing the segmentation accuracy without any extra work done by the end-user, at the cost of higher memory usage and potentially slower inference speed.

\begin{table}[ht]
\centering
\begin{tabular}{l|c|c}
\hline
\textbf{Increase brightness} & 0.721 & +0.009 \\ \hline
\textbf{Decrease brightness} & 0.725 & +0.013 \\ \hline
Grayscale & 0.707 & -0.005 \\ \hline
\textbf{Reduce bits to $3$} & 0.717 & +0.005 \\ \hline
\textbf{Make sharp} & 0.718 & +0.006 \\ \hline
\textbf{Gaussian blur} & 0.731 & +0.019 \\ \hline
\textbf{Rotate right $45\deg$}\dag & 0.723 & +0.011 \\ \hline
Translate right +100 px & 0.675 & -0.037 \\ \hline
\textbf{Zoom out $0.5 \times$} & 0.715 & +0.003 \\ \hline
\textbf{Zoom in $1.5 \times$} & 0.727 & +0.015 \\ \hline
\textbf{Shear right by $20$}\dag & 0.730 & +0.018 \\ \hline
Crop mask region & 0.704 & -0.008 \\ \hline
\end{tabular}
\caption{In-memory augmentations in their individual effect on the overall segmentation quality on LVOS dataset. Only transformations named in \textbf{bold} were considered for experiments. For transformation with \dag the equivalent symmetrical transform was used as well. A total of 11 augmentations were used for the experiments in Fig. \ref{fig:scaling}.}
\label{tab:augs}
\end{table}

\paragraph{Utility of Permanent Memory.}
We further demonstrate the capabilities of our introduced permanent memory module by disabling updates to the temporary memory in \modelname{}, effectively keeping it empty and frozen throughout the inference. We observe that for LVOS dataset (Fig. \ref{fig:scaling}) this results in an \textit{increase} in the overall segmentation quality, both with (red) and without (purple) using in-memory augmentations. This shows that for certain types of videos, especially when predicted masks are prone to have errors, the best strategy is to not use them at all, and very few high-quality references in the memory result in higher segmentation quality than dozens, potentially hundreds, but containing errors. Temporary memory plays has an important role in \modelname{} architecture, allowing it to adapt better to changes in the target appearance, but through our experiments we show that for some videos it can be safely disabled (for example, if the target object's appearance stays relatively consistent throughout the video), leading to higher inference speed and lower memory footprint.

\clearpage
\label{sec:alg}

\newcommand{\keys}{\mathbf{K}}
\newcommand{\masks}{\mathbf{M}}
\newcommand{\prevcand}{\mathbf{PC}}
\newcommand{\composite}{\mathbb{K}}
\newcommand{\compositeKey}{\hat{k}}
\newcommand{\norm}[1]{\lvert #1 \rvert}
\newcommand{\chosen}{\mathbf{CC}}
\newcommand{\chosenKeys}{\mathbf{C\hat{K}}}
\newcommand{\dis}{\neg\mathbf{S}}
\newcommand{\disMin}{\neg S_{min}}
\newcommand{\disMem}{\neg \mathbf{S}_{\keys}}

\begin{algorithm*}[!htb]
Here is the pseudo-code for the annotation candidate selection algorithm. The $\odot$ is a pointwise multiplication operation. Symbol $[\:]$ denotes an empty list, and symbols $\mathbf{S}$ and $\dis$ are used to denote similarity and dissimilarity correspondingly (the negation symbol $\neg$ is used as a visual cue) 
    \caption{select-next-candidates($\keys$, $\masks$, k, $\prevcand$, $\alpha=0.5$, $\beta=9$)}
    \begin{algorithmic}[1]
    \Require $\keys$: list of ``key" feature maps for all frames of the video
    \Require $\masks$: list of masks for each frame (predicted or user-provided)
    \Require $k$: number of candidate frames to select
    \Require $\prevcand$: list of previously chosen candidate indices (default is [0])
    \Require $\alpha$: weight of mask regions (default is 0.5), $\alpha \in [0..1]$
    \Require $\beta$: minimum number of pixels for a valid mask, to explicitly filter out frames without the target object or where it is too small (default is 9px)
    
    \Ensure $\prevcand$ not empty, $0 \leq \alpha \leq 1.0$, $k > 0$
    
    \Function{select-next-candidates}{$\keys$, $\masks$, $k$, $\prevcand$, $\alpha$, $\beta$}
    \State $\composite \gets [\:],$  $N \gets$ $\norm{\keys} $\Comment{Composite keys, Number of frames}

    \For{$i$ in $[0, N-1]$}
        \State $\compositeKey \gets \keys[i] \odot \masks[i] \cdot \alpha + \keys[i] \cdot (1 - \alpha)$ \Comment{$\compositeKey$ is a ``composite" key}
        \State \Comment{Equivalent to alpha-blending operation}
        \State $\composite$.add\_to\_end($\compositeKey$) 
    \EndFor
    
    \State $\chosen \gets \prevcand$ \Comment{Chosen candidates, initialize with previous candidates}
    \State $\chosenKeys \gets \left[\composite[i] \mid i \in \prevcand \right]$ \Comment{Chosen candidates composite keys}
    
    \For{$i$ in [0, $k$]}
        \State $\dis \gets [\:]$ \Comment{Dissimilarities between candidates and other frames}
        \For{$j$ in [0, $N$]}
            \If{$\norm{\masks[i] > 0} < \beta$}
                \Comment{Mask empty or too small, ignore}
                \State $\disMin \gets 0$ \Comment{Minimum dissimilarity of frame $i$ to all in $\chosen$}
            \Else
                \State $\disMem \gets [\:]$ \Comment{Dissimilarities of $i \rightarrow j, \forall j \in \chosen$}
                \For{$j$ in $\chosen$}
                    \State $S_{j \rightarrow i} \gets$ similarity($\chosenKeys[j]$, $\composite[i]$)
                    \State $S_{i \rightarrow j} \gets$ similarity($\composite[i]$, $\chosenKeys[j]$)
    
                    \State $\dis_{cycle} \gets (S_{j \rightarrow i} - S_{i \rightarrow j})$ \Comment{Pixel-wise cycle dissimilarity}
                    \State $\neg S_{cycle} \gets \frac{\sum{\max(0, \dis_{cycle})}}{\norm{\dis_{cycle}}}$ \Comment{Only non-negative mappings}
                    \State $\disMem$.add\_to\_end($\neg S_{cycle}$) 
                \EndFor
                \State $\disMin \gets \min(\disMem)$
            \EndIf
            \State $\dis$ .add\_to\_end($\disMin$)
        \EndFor
        \State $c \gets \text{argmax}(\dis)$ \Comment{New selected candidate}
        \State $\chosen$.add\_to\_end($c$)
        \State $\chosenKeys$.add\_to\_end($\composite[c]$)
    \EndFor
    \State \Return $\left[\chosen[i] \mid i \geq \norm{\prevcand}\right]$ \Comment{Return new candidates, from index $\norm{\prevcand}$}
    \EndFunction
    \end{algorithmic}
\label{alg:alg_candidates}
\end{algorithm*}